\documentclass[10pt,twocolumn,letterpaper]{article}

\usepackage[pagenumbers]{cvpr}      
\usepackage{marvosym}

\usepackage{array}  
\newcolumntype{P}[1]{>{\centering\arraybackslash}p{#1}} 

\definecolor{cvprblue}{rgb}{0.21,0.49,0.74}
\usepackage[pagebackref,breaklinks,colorlinks,allcolors=cvprblue]{hyperref}
\usepackage{soul}
\usepackage{booktabs}
\usepackage{multirow}
\usepackage{pifont}       

\usepackage{bbding}       
\usepackage{fontawesome}  

\newcommand{\myparagraph}[1]{{\vspace{.2em} \noindent \bf #1}}

\title{LoVoRA: Text-guided and Mask-free Video Object Removal and Addition with Learnable Object-aware Localization}

\author{
Zhihan Xiao\textsuperscript{1} \quad
Lin Liu\textsuperscript{2}\textsuperscript{\Letter}\thanks{Project Leader.} \quad
Yixin Gao\textsuperscript{3} \quad
Xiaopeng Zhang\textsuperscript{2} \\
Haoxuan Che\textsuperscript{2} \quad
Songping Mai\textsuperscript{1}\textsuperscript{\Letter} \quad
Qi Tian\textsuperscript{2} \\
$^1$ Tsinghua University \quad
$^2$ Huawei Inc. \quad
$^3$ University of Science and Technology of China
}

\begin{document}
\maketitle

\renewcommand{\thefootnote}{\Letter}
\footnotetext[0]{
Corresponding authors: \texttt{ll0825@mail.ustc.edu.cn}
\texttt{mai.songping@sz.tsinghua.edu.cn}, 
}
\renewcommand{\thefootnote}{\arabic{footnote}}

\begin{figure*}[t]
    \centering
    \includegraphics[width=\linewidth]{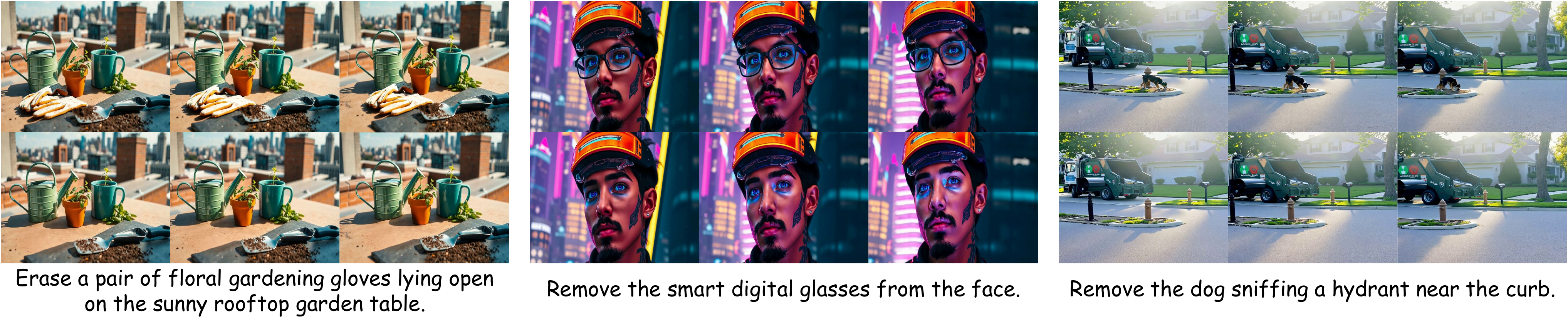}
    \includegraphics[width=\linewidth]{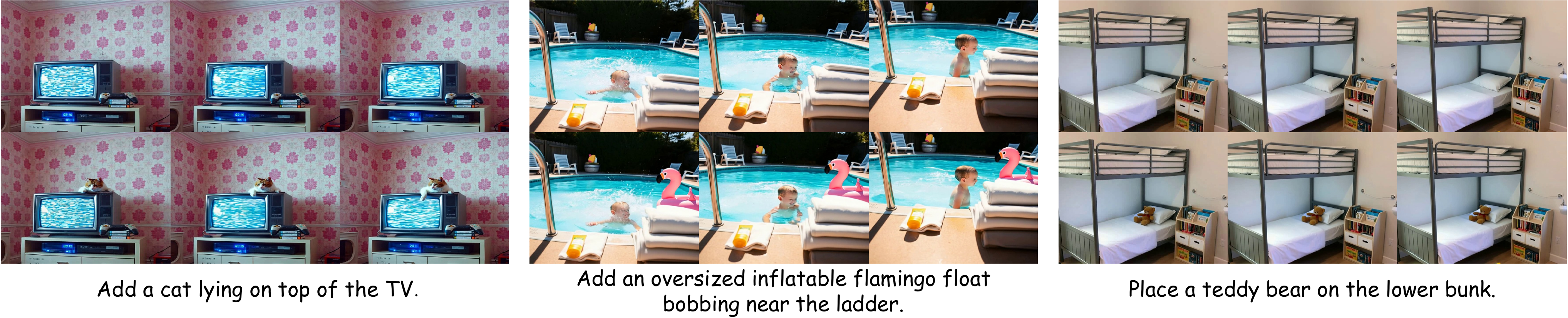}
    \caption{Removal and addition samples from our LoVoRA dataset. The order from top to bottom is: source video, edited video, and editing instruction.}
    \label{fig:pipeline_overview}
\end{figure*}

\begin{abstract}
Text-guided video editing, particularly for object removal and addition, remains a challenging task due to the need for precise spatial and temporal consistency. Existing methods often rely on auxiliary masks or reference images for editing guidance, which limits their scalability and generalization. To address these issues, we propose LoVoRA, a novel framework for mask-free video object removal and addition using object-aware localization mechanism. Our approach utilizes a unique dataset construction pipeline that integrates image-to-video translation, optical flow-based mask propagation, and video inpainting, enabling temporally consistent edits. The core innovation of LoVoRA is its learnable object-aware localization mechanism, which provides dense spatio-temporal supervision for both object insertion and removal tasks. By leveraging a Diffusion Mask Predictor, LoVoRA achieves end-to-end video editing without requiring external control signals during inference. Extensive experiments and human evaluation demonstrate the effectiveness and high-quality performance of LoVoRA. \href{https://cz-5f.github.io/LoVoRA.github.io/} {https://cz-5f.github.io/LoVoRA.github.io/}
\end{abstract}

\section{Introduction}
\label{sec:intro}
Video object removal and addition are two essential forms of video editing that enable fine-grained scene manipulation at the object level. 
Achieving both requires accurate spatial localization, temporal coherence, and semantic understanding while maintaining the global consistency of motion, lighting, and background.

Recent advances in diffusion-based video generation models have dramatically improved the realism and controllability of video editing~\cite{wang2023videocomposer,zhang2024avid,bian2025videopainter,zi2025minimax,wan2025unipaint,lucyedit,ditto,icve,senoritia,ku2024anyv2v,insv2v,lgvi2023,omniinsert}.
Most existing methods, however, rely on explicit auxiliary inputs such as masks, reference images, or control signals to guide the edit region. For instance, VideoComposer~\cite{wang2023videocomposer} and AVID~\cite{zhang2024avid} perform mask-guided inpainting conditioned on textual prompts, while GetInVideo~\cite{zhuang2025getinvideo} and VideoAnyDoor~\cite{tu2025videoanydoor} further introduce reference objects to control spatial position and motion trajectory.
Although effective, these approaches exhibit two critical limitations:
(\textit{i}) their editing quality heavily relies on the precision of user-provided masks or reference frames, and often requires tedious user feedback or manual adjustments, making them impractical for real-world use; and (\textit{ii}) most existing models are designed for a single editing type, restricting their applicability in open-ended video editing. 

To address the diversity of editing objectives, multi-task frameworks such as MTV-Inpaint~\cite{Mtv-inpaint}, VideoPainter~\cite{bian2025videopainter}, and O-DisCo-Edit~\cite{chen2025disco} have emerged, supporting multiple tasks like addition, removal, or style modification in a unified manner.
However, these methods still depend on explicit masks during inference. In practice, a fully end-to-end, mask-free video editing framework that can accurately localize and modify objects purely from textual instructions remains largely unexplored.

In this paper, we present LoVoRA, a text-guided and mask-free video object removal and addition framework that operates in a fully end-to-end manner.
LoVoRA eliminates the reliance on explicit spatial masks or external control signals, instead learning to implicitly identify and modify object regions guided solely by text prompts.
To achieve this, we introduce two complementary innovations:  (\textit{i}) A high-quality video editing dataset specifically constructed for object-level addition and removal, providing precise motion-aware masks and optical-flow-based supervision; and (\textit{ii}) A learnable object-aware localization mechanism integrated into a video editing backbone, allowing the model to infer where and how to edit directly from multimodal context.

Our dataset is built from the NHR-Edit~\cite{nhredit} image editing corpus through a multi-stage synthesis pipeline involving image-to-video translation, optical flow estimation, and mask propagation. The resulting data pairs provide temporally consistent source-target video sequences, text instructions, and dense motion-mask annotations. Such high-quality supervision is particularly crucial for mask-free editing, where the model must implicitly learn both spatial boundaries and temporal transitions without relying on hand-crafted annotations. On top of this, LoVoRA employs a Diffusion Mask Predictor that learns soft spatio-temporal masks in a self-supervised manner, guided by mask-aware training objectives. This component helps the model focus its editing capacity on semantically meaningful regions without requiring any masks during inference.

Detailed studies further demonstrate that our data synthesis strategy produces higher temporal consistency and more reliable motion supervision compared with existing video editing dataset generation methods. 
Extensive experiments on public and self-constructed benchmarks demonstrate the effectiveness of LoVoRA.
Compared with very recent video editing frameworks such as Ditto~\cite{ditto} and LucyEdit~\cite{lucyedit}, our method achieves superior performance in prompt following and edit quality. Qualitative results show that LoVoRA can seamlessly remove or insert objects while preserving background structure, lighting, and motion continuity even in complex scenes with occlusion or fast motion. 

In summary, our key contributions are as follows:\begin{itemize}
\item A novel end-to-end framework, LoVoRA, that performs both text-guided video object removal and addition without requiring any masks or external control signals at inference time.
\item A high-quality video editing dataset featuring temporally consistent mask flow supervision, optical-flow guidance, and semantically aligned textual instructions for object-aware local editing. Our dataset will be open-sourced.
\item A mask-aware training strategy with a Diffusion Mask Predictor that enables implicit, learnable localization of edited regions, achieving spatial precision and temporal stability superior to existing methods.
\end{itemize}
\section{Related Works}
\label{sec:related}
\subsection{Video object removal and addition}
Video object removal and addition are two fundamental tasks of video editing that modify scene content at the object level~\cite{sun2024diffusion,wang2023videocomposer,Mtv-inpaint,bian2025videopainter,zi2025minimax,zi2025cococo,zhang2024avid,wan2025unipaint,zhuang2025getinvideo,omniinsert,expressedit}. Object addition aims to insert novel entities that are semantically and visually consistent with the surrounding context, while object removal reconstructs plausible backgrounds after erasing unwanted regions.

\myparagraph{Single-task model.}
The success of video generation foundation models has spurred the development of frameworks focusing on either object addition or removal. Object addition methods can be categorized by their conditioning inputs into two types: (i) text+mask, and (ii) reference+mask. For the first type, diffusion-based models such as VideoComposer~\cite{wang2023videocomposer} and AVID~\cite{zhang2024avid} adopt mask-guided inpainting conditioned on textual prompts, where the mask defines the spatial editing region and the text controls semantic content. CoCoCo~\cite{zi2025cococo} and UniPaint~\cite{wan2025unipaint} further enhance this paradigm through improved text-video alignment and space-time modeling for more consistent and controllable inpainting. For the second type, GetInVideo~\cite{zhuang2025getinvideo} and VideoAnyDoor~\cite{tu2025videoanydoor} employ reference images or object trajectories in addition to masks to guide object insertion with fine-grained motion control and temporal coherence. 
In parallel, most object removal methods only rely on explicit mask inputs to locate the target object to be erased~\cite{zhou2023propainter,zi2025minimax,li2025diffueraser,rose,wei2025omnieraser}. For example, MiniMax-Remover~\cite{zi2025minimax} introduces a two-stage framework that removes cross-attention and applies minimax optimization to resist bad noise, enabling fast and CFG-free video object removal. ROSE~\cite{rose} introduces a framework that leverages synthetic 3D-rendered paired data and explicit side-effect mask supervision to jointly remove objects and their associated effects such as shadows and reflections.

\myparagraph{Multi-task model.} To meet the growing diversity of user creation demands, recent studies have explored unified frameworks that support multiple video editing objectives within a single model. MTV-Inpaint~\cite{Mtv-inpaint} unifies object addition and removal within a multi-task video inpainting framework using a dual-branch spatial attention design for consistent foreground synthesis and background completion. VideoPainter~\cite{bian2025videopainter} achieves both object addition and removal through a dual-branch diffusion transformer with a lightweight context encoder and ID-resampling strategy, enabling controllable and identity-consistent editing for videos of any length. Beyond these, some works further generalize toward broader editing capabilities such as style or light transfer~\cite{vace,chen2025disco,ju2025editverse,lucyedit,liang2025omniv2v,ye2025unic}. For instance, VACE~\cite{vace} introduces a unified editing backbone that integrates multiple visual conditions for controllable video generation, while O-DisCo-Edit~\cite{chen2025disco} introduces a unified noise-based control signal that encapsulates diverse editing cues, effectively supporting object addition, removal, and other complex video edits within a single framework. However, most existing frameworks rely on additional signals (\eg masks and reference images) during inference, and their editing quality often hinges on the precision of these auxiliary conditions. In contrast, our method avoids the dependence on external inputs through a carefully designed end-to-end framework and achieves superior background consistency and accurate localization of edited regions.




\subsection{Text-guided video editing datasets} 
Recently, several high-quality and large-scale synthetic datasets have significantly advanced the development of text-guided video editing. Although VIVID-10M~\cite{hu2024vivid} offers a large set of videos paired with textual instructions and mask annotations, it does not include edited videos as ground truth, which limits its applicability for training video editing models. InsViE-1M~\cite{insvie} utilizes a two-stage pipeline that edits and screens the first frame using an image editor and GPT-4o, then propagates it into a full video through I2V generation with additional filtering by GPT-4o and optical-flow consistency. Senorita-2M~\cite{senoritia} constructs its dataset through multiple expert models, while Ditto-1M~\cite{ditto} builds its dataset via an automated three-stage pipeline of video filtering, image- and depth-guided generation, and VLM-based quality enhancement. However, these datasets still provide limited coverage of object addition and removal scenarios and often suffer from inconsistent backgrounds caused by frame-wise generation, making them less suitable for training object addition and removal editing models.

\section{Dataset Construction}
\label{sec:dataset}

\subsection{Motivation and Overview}
Existing instruction-based video editing datasets~\cite{insvie,ditto,senoritia,lucyedit} have significantly advanced text-guided video manipulation. However, they still suffer from several limitations: (\textit{i}) low spatial and temporal resolution, (\textit{ii}) inconsistent backgrounds caused by frame-wise generation, and (\textit{iii}) limited coverage of object addition and removal scenarios. 
To overcome these issues, we construct a high-quality dataset specialized for object-level addition and erasure in videos, which serves as the foundation of our proposed framework. Our dataset is built upon high-fidelity image editing dataset NHR-Edit~\cite{nhredit} and synthesized into temporally consistent video editing sequences via a multi-stage pipeline (see \cref{fig:pipeline_overview}). 
Compared to prior works, our dataset provides \textbf{(a)} better background consistency, \textbf{(b)} accurate object-level spatio-temporal masks, and \textbf{(c)} robust text instructions aligned with editing semantics. Moreover, it provides dense motion-mask supervision that enables training and evaluation of object-aware edit localization in LoVoRA.

\begin{figure*}[t]
    \centering
    \setlength{\abovecaptionskip}{4pt}
    \includegraphics[width=\linewidth]{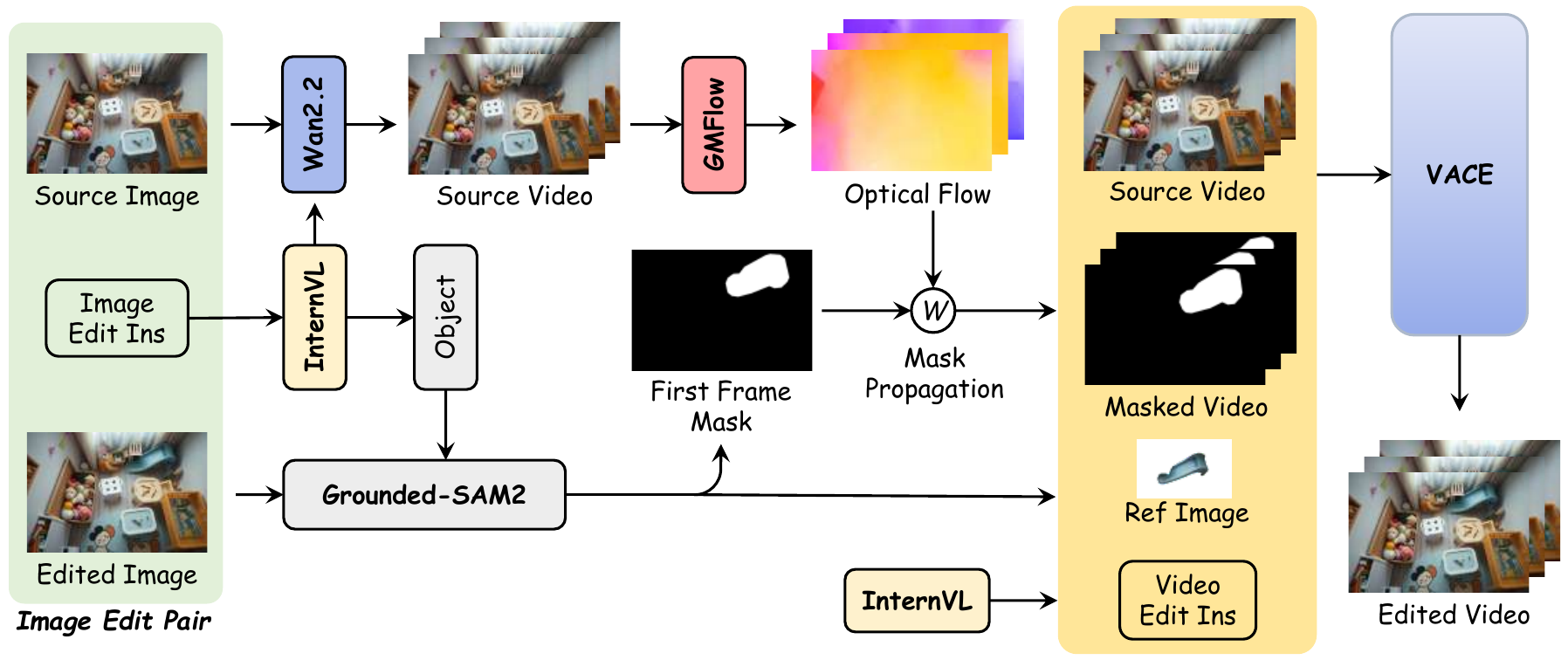}
    \caption{Overview of LoVoRA dataset construction pipeline. Starting from high-quality image editing pairs, we synthesize instruction-based video editing data through five: I2V translation, mask generation, optical flow estimation, mask propagation, and video inpainting.}
    \label{fig:pipeline_overview}
\end{figure*}

\subsection{Pipeline Overview}
Given an image editing pair $(I_s, I_t)$ and an instruction $\mathbf{p}$ describing the editing operation (e.g., \textit{``add a plastic slide in the corner''}), we synthesize a corresponding video editing pair $(V_s, V_t)$ through five major stages:

\vspace{0.5em}
\noindent\textbf{Stage 1: Image-to-Video Generation.}~We first synthesize a source video $V_s = \{I_t^1, I_t^2, \dots, I_t^T\}$ from the source image $I_s$ using a text-conditioned image-to-video model, Wan2.2~\cite{wan22}. This model generates temporally coherent frames preserving the spatial layout and background consistency:
\begin{equation}
V_s = \Phi_{\text{I2V}}(I_s, \mathbf{p}_{\text{scene}}),
\end{equation}
where $\Phi_{\text{I2V}}$ denotes the video synthesis function, and $\mathbf{p}_{\text{scene}}$ is the scene description extracted from $\mathbf{p}$ using InternVL3~\cite{InternVL3}.

\vspace{0.5em}
\noindent\textbf{Stage 2: Mask Generation.}~To accurately localize the edited region, we generate a single object-level mask on the first frame of the source or target image, depending on the editing type. 
Given the instruction $\mathbf{p}$, we first employ Grounding DINO~\cite{grounding} to detect the referred object and obtain its bounding box $b$. 
This bounding box is then fed into SAM2~\cite{sam2} to produce a precise binary segmentation mask $m$ corresponding to the edited object. 
For an editing pair $(I_s, I_t)$, the initial mask $M_1$ is defined as:
\begin{equation}
M_1 =
\begin{cases}
m^{(t)}, & \text{if addition task}, \\[4pt]
m^{(s)}, & \text{if deletion task},
\end{cases}
\end{equation}
where $m^{(t)}$ and $m^{(s)}$ are masks extracted from the target and source images, respectively. 
We further refine $M_1$ via morphological smoothing. This ensures that the generated mask accurately captures the object boundary and provides a reliable initialization for subsequent mask propagation.

\vspace{0.5em}
\noindent\textbf{Stage 3: Optical Flow Estimation.}~We compute dense optical flow $\mathcal{F}_{t \rightarrow t+1}$ between consecutive frames of $V_s$ using GMFlow~\cite{gmflow}. This provides pixel-wise motion guidance for constructing the mask flow field, which jointly encodes object movement and spatial edit regions across time.

\vspace{0.5em}
\noindent\textbf{Stage 4: Mask Propagation.}~The initial mask $M_1$ is temporally extended through the optical flow field $\mathcal{F}_{t \rightarrow t+1}$ to form a continuous mask flow $\{M_t\}_{t=1}^{T}$:
\begin{equation}
M_{t+1} = \mathcal{W}(M_t, \mathcal{F}_{t \rightarrow t+1}),
\end{equation}
where $\mathcal{W}$ denotes a backward warping operator. 
Unlike discrete per-frame masks, $M_t$ represents a temporally coherent field that evolves smoothly with motion, providing dense spatio-temporal supervision for LoVoRA.
We further apply bidirectional flow consistency to handle occlusion and maintain precise boundaries.

\vspace{0.5em}
\noindent\textbf{Stage 5: Video Inpainting.}~Given the derived mask flow $\{M\}$, source video $V_s$, and editing instruction $\mathbf{p}$, we synthesize the edited video $V_t$ using VACE~\cite{vace}.
The inpainting function
\begin{equation}
V_t = \Phi_{\text{VACE}}(V_s, \{M\}, \mathbf{p})
\end{equation}
fills in or removes content inside masked regions while preserving scene layout and motion continuity. 
By conditioning on both the instruction and the surrounding context, VACE produces realistic and temporally consistent results, leading to high-quality video pairs $(V_s, V_t)$ for addition and erasure tasks.

\subsection{Instruction Generation and Filtering}
The textual instruction $\mathbf{p}$ and object prompts are generated using InternVL3~\cite{InternVL3}, ensuring accurate alignment between visual edits and linguistic expressions. To maintain data quality, we apply automatic filtering:
\begin{align}
\text{keep}(V_s, V_t) = 
&~[\text{area}(M) \in [\alpha_{\min}, \alpha_{\max}]] \nonumber\\
&\wedge [\text{flow\_mag}(V_s) \in [\beta_{\min}, \beta_{\max}]],
\end{align}
where thresholds $(\alpha_{\min}, \alpha_{\max}, \beta_{\min}, \beta_{\max})$ empirically ensure sufficient motion and region significance.

\begin{table*}[t]
\small
\caption{Comparison of video editing datasets. Our dataset emphasizes high-resolution, temporally consistent object manipulation with optical flow guided mask propagation. We also report VLM evaluation results using MiniCPM-V2.6. PF denotes  Prompt Following and EQ denotes Edit Quality.}
\centering
\begin{tabular}{l|ccc|cc|p{5.8cm}}
\toprule
Dataset & Resolution & FPS & Frames & PF & EQ & Generation Basis \\
\midrule
InsV2V~\cite{insv2v} & $256\times256$ &  -- & 16 & -- & -- & Modified Prompt-to-Prompt method \\

ICVE-SFT~\cite{icve}& $480\times480$-$576\times1024$  & -- & --  & -- & -- & Source object removal and inpainting \\
Senorita-2M~\cite{senoritia} &  $336\times592$-$1120\times1984$ & 8 & 33–64 & 3.533 & 3.883 & Source object removal and inpainting \\
InsViE-1M~\cite{insvie} & $1024\times576$ & 7 & 25 & 3.133 & 3.667 & Video inversion and reconstruction guidance \\
Ditto~\cite{ditto} & $1280\times720$ & 20 & 101 & \textbf{4.417} & 4.733 & Depth-guided generation from source video \\
\textbf{Ours} & \textbf{720p} & \textbf{16–24} & \textbf{81–121} & 4.375 & \textbf{4.850} & \textbf{Optical flow guided mask propagation} \\
\bottomrule
\end{tabular}
\label{tab:dataset_comparison}
\end{table*}

\subsection{Dataset Statistics and Comparison}
Our dataset contains 8K video editing pairs with resolutions up to 720p and frame rates ranging from 16 to 24~fps, higher than those in previous datasets (see \cref{tab:dataset_comparison}). Each video is accompanied by a rich editing instruction, a temporally consistent mask sequence, and an optical flow field, providing fine-grained supervision for both object addition and removal tasks. 

To further evaluate the editing quality, we employ a Vision-Language Model (VLM), MiniCPM-V2.6~\cite{minicpm}, to assess the results in terms of Prompt Following (PF) and Edit Quality (EQ). Both metrics are scored on a scale from 0 to 5, with higher values indicating better performance. PF measures how well the edited video aligns with the given textual instruction, while EQ evaluates the perceptual realism and temporal consistency of the edit. For fairness and diversity, we randomly sample a subset of cases from each dataset for evaluation. As shown in \cref{tab:dataset_comparison}, our dataset achieves competitive PF and the highest EQ score among all compared datasets, demonstrating superior visual fidelity and coherence across frames. These results highlight the effectiveness of our optical flow guided mask propagation in producing temporally stable and semantically accurate video edits.

\begin{figure*}[t]
\centering
\setlength{\abovecaptionskip}{2pt}
\includegraphics[width=1\linewidth]{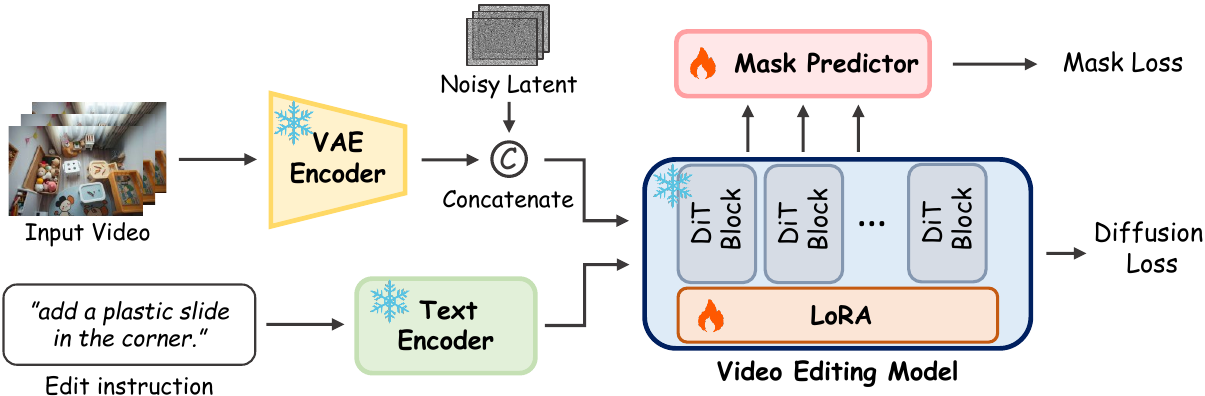}
\caption{Overall architecture. The input video is encoded by a spatio-temporal VAE to produce latents. Encoded latents are channel-concatenated with noisy target latents and processed by a DiT backbone to predict the rectified-flow velocity field. A Diffusion Mask Predictor reads selected DiT token features and predicts a spatio-temporal diff mask used during training.}
\label{fig:model_overview}
\end{figure*}

\section{LoVoRA Model}
\label{sec:LoVoRA Model}
We introduce LoVoRA, a text-guided model that supports both video object addition and removal in an end-to-end manner. Unlike most previous approaches~\cite{wang2023videocomposer,wan2025unipaint,zhuang2025getinvideo,tu2025videoanydoor,zhang2024avid,Mtv-inpaint,bian2025videopainter} that rely on explicit auxiliary inputs such as masks, reference frames, or control signals, LoVoRA performs editing solely based on textual instructions during inference. To achieve this, we design a learnable object-aware edit localization mechanism that enables the model to implicitly identify regions requiring modification while preserving non-edited content. Subsequent sections detail our model architecture (Section~\ref{sec:architecture}) and our proposed learnable object-aware edit localization mechanism (Section~\ref{sec:mask-predictor}).


\vspace{0.5em}
\subsection{Architecture}
\label{sec:architecture}
As shown in ~\cref{fig:model_overview}, LoVoRA model builds on a spatio-temporal rectified-flow video editing backbone~\cite{lucyedit}.
The input video $V \in \mathbb{R}^{3 \times F \times H \times W}$ is first encoded into a latent representation $Z \in \mathbb{R}^{C \times F_p \times H_p \times W_p}$ through a spatio-temporal VAE, while the text instruction $\mathbf{p}$ is transformed into text embeddings via a text encoder. A noisy latent $\tilde{Z} \in \mathbb{R}^{C \times F_p \times H_p \times W_p}$ is then sampled and concatenated with the encoded video latent along the channel dimension:
\begin{equation}
    Z_\text{cat} = \text{Concat}(Z, \tilde{Z}) \in \mathbb{R}^{2C \times F_p \times H_p \times W_p}.
\end{equation}
Subsequently, the fused feature $Z_\text{cat}$ is fed into a 3D DiT-based backbone, where the text embeddings are injected through cross-attention layers to guide the denoising process according to the editing instruction. This backbone predicts the velocity field that transports the noisy latent toward the clean latent manifold under the rectified-flow formulation, progressively generating the edited video $\hat{V}$.



\vspace{0.5em}
\subsection{Learnable Object-aware Edit Localization}
\label{sec:mask-predictor}
A key challenge in video editing is accurately identifying \emph{where to apply edits}.
Relying solely on textual guidance often causes global over-editing or incomplete local edits. To address this, we introduce a lightweight Diffusion Mask Predictor, which learns to localize edit-relevant regions in a self-supervised manner. 
This predictor guides the model to edit the relevant regions and avoid unnecessary changes, leading to better spatial accuracy and temporal consistency.

Given intermediate latent tokens $X \in \mathbb{R}^{B \times L \times D}$ from the DiT model, we first apply a projection MLP:
\begin{equation}
    M' = \phi(X) \in \mathbb{R}^{B \times L \times 1},
\end{equation}
where $\phi(\cdot)$ is a two-layer feed-forward network with GELU activations. 

We then reshape $M'$ back to the spatio-temporal grid $(F_p, H_p, W_p)$ and upsample via trilinear interpolation to match the original video resolution:
\begin{equation}
    M = \text{Interp}(M', \, (F, H, W)).
\end{equation}
The resulting mask $M \in \mathbb{R}^{B \times 1 \times F \times H \times W}$ serves as a soft edit localization signal learned automatically from the dataset, without explicit supervision at inference time.

The overall loss function includes three parts. First, the base diffusion loss minimizes mean squared error on velocity prediction:
\begin{equation}
    \mathcal{L}_{\text{diff}} = \| v_\theta(Z_\text{cat}, P) - v^{*} \|^{2}.
\end{equation}

To emphasize correctness in edited regions while suppressing trivial background matching, we introduce a mask-weighted loss:
\begin{equation}
    \mathcal{L}_{\text{mask}} = \| (v_\theta - v^{*}) \odot M^{*} \|^{2}.
\end{equation}

Additionally, we use the ground-truth difference masks $M^{*}$ to supervise the estimated soft mask $M$ through  binary cross-entropy (BCE) loss: 
\begin{equation}
    \mathcal{L}_{\text{pred}} = \text{BCE}(M, M^{*}).
\end{equation}
It is important to note that these ground-truth masks $M^{*}$ are only used during training to guide the model in learning spatial localization of edits and are not required during inference.

Finally, the overall training objective is:
\begin{equation}
    \mathcal{L} = \mathcal{L}_{\text{diff}} + \lambda_1 \mathcal{L}_{\text{mask}} + \lambda_2 \mathcal{L}_{\text{pred}}.
\end{equation}

After training, the mask predictor is not included in the inference stage. The proposed learnable object-aware edit localization strategy enables LoVoRA to implicitly learn spatially precise and semantically guided editing without any auxiliary inputs during inference.
\textbf{\begin{table*}[t]
\centering
\caption{Quantitative comparison on the object removal task. 
We report results across three evaluation dimensions: Text Alignment, Video Quality, and VLM-based evaluation. PF denotes Prompt Following and EQ denotes Edit Quality. 
Best results are bolded.}
\label{tab:removal}
\resizebox{\linewidth}{!}{
\begin{tabular}{c|c|cc|ccccc|cc}
\toprule
\multirow{2}{*}{\textbf{Removal}} &
\multirow{2}{*}{\textbf{Ref. Img.}} &
\multicolumn{2}{c|}{\textbf{Text Alignment}$\uparrow$} & 
\multicolumn{5}{c|}{\textbf{Video Quality}$\uparrow$} & 
\multicolumn{2}{c}{\textbf{VLM}$\uparrow$} \\
\cmidrule(lr){3-4} \cmidrule(lr){5-9} \cmidrule(lr){10-11}
& & Frame & Video & Subj.\ Cons. & Backg.\ Cons. & Motion & Aesthetic & Imaging & PF & EQ \\
\midrule
Senorita~\cite{senoritia}  & \Checkmark & 19.39 & 13.64  & \textbf{0.9424} & 0.9256  & 0.9824 & 0.5338 & 0.5557 & 4.117 & 4.314 \\
InsV2V~\cite{insv2v}       & \XSolidBrush & 20.30 & 15.72 & 0.9119 & 0.9067 & 0.9873 & 0.5432 & 0.4483 & 3.824 & 4.176 \\
LGVI~\cite{lgvi2023}       & \XSolidBrush & 20.24 & 13.22 & 0.9223 & 0.9083 & 0.9704 & 0.4661 & 0.5300 & 4.412 & 4.588 \\
Ditto~\cite{ditto}         & \XSolidBrush & \textbf{20.80} & \textbf{15.96} & 0.9330 & 0.9128 & 0.9876 & \textbf{0.5987} & 0.6469 & 3.882 & 4.471 \\
LucyEdit~\cite{lucyedit}   & \XSolidBrush & 20.28 & 15.14 & 0.9013 & 0.9120 & 0.9885 & 0.5564 & \textbf{0.6606} & 3.824 & 4.647 \\
\textbf{Ours}              & \XSolidBrush & 20.68 & 14.62 & 0.9355 & \textbf{0.9286} & \textbf{0.9886} & 0.5683 & 0.6402 & \textbf{4.608} & \textbf{4.882} \\
\bottomrule
\end{tabular}
}
\end{table*}
}
\begin{table*}[t]
\centering
\caption{Quantitative comparison on the object addition task.
We report results across three evaluation dimensions: Text Alignment, Video Quality, and VLM-based evaluation. PF denotes Prompt Following and EQ denotes Edit Quality. Best results are bolded.}
\label{tab:addition}
\resizebox{\linewidth}{!}{
\begin{tabular}{c|c|cc|ccccc|cc}
\toprule
\multirow{2}{*}{\textbf{Addition}} &
\multirow{2}{*}{\textbf{Ref. Img.}} &
\multicolumn{2}{c|}{\textbf{Text Alignment}$\uparrow$} &
\multicolumn{5}{c|}{\textbf{Video Quality}$\uparrow$} &
\multicolumn{2}{c}{\textbf{VLM}$\uparrow$} \\
\cmidrule(lr){3-4} \cmidrule(lr){5-9} \cmidrule(lr){10-11}
&  & Frame & Video & Subj.\ Cons. & Backg.\ Cons. & Motion & Aesthetic & Imaging & PF & EQ \\
\midrule
Senorita~\cite{senoritia} & \Checkmark & 22.64 & \textbf{18.41}  & 0.9215  & 0.9186 & 0.9827 & 0.5380 & 0.6090 & 3.158& 3.421\\
InsV2V~\cite{insv2v} & \XSolidBrush & 22.81 & 17.15 & 0.8717 & 0.8893 & 0.9846 & 0.5519 & 0.4897 & 3.158 & 3.769 \\
LGVI~\cite{lgvi2023}   & \XSolidBrush  & 17.70 & 10.56 & 0.8939 & 0.8309 & 0.9695 & 0.4049 & 0.5087 & 1.579 & 2.982 \\
Ditto~\cite{ditto}  & 
\XSolidBrush & 19.08 & 13.20 & 0.9238 & 0.8965 & 0.9858 & \textbf{0.5567}& 0.6208 & 2.807 & 3.684 \\
LucyEdit~\cite{lucyedit} & 
\XSolidBrush & 22.75 & 17.01 & 0.9188 & \textbf{0.9317}& 0.9878 & 0.5478 & 0.6326 & 3.070 & 4.298 \\
\textbf{Ours} & \XSolidBrush & \textbf{23.00} & 17.50 & \textbf{0.9274} & 0.9079& \textbf{0.9889} & 0.5504& \textbf{0.6420} & \textbf{3.421} & \textbf{4.386} \\
\bottomrule
\end{tabular}}
\end{table*}

\begin{figure*}[t]
    \centering
    \setlength{\abovecaptionskip}{2pt}
    \includegraphics[width=\linewidth]{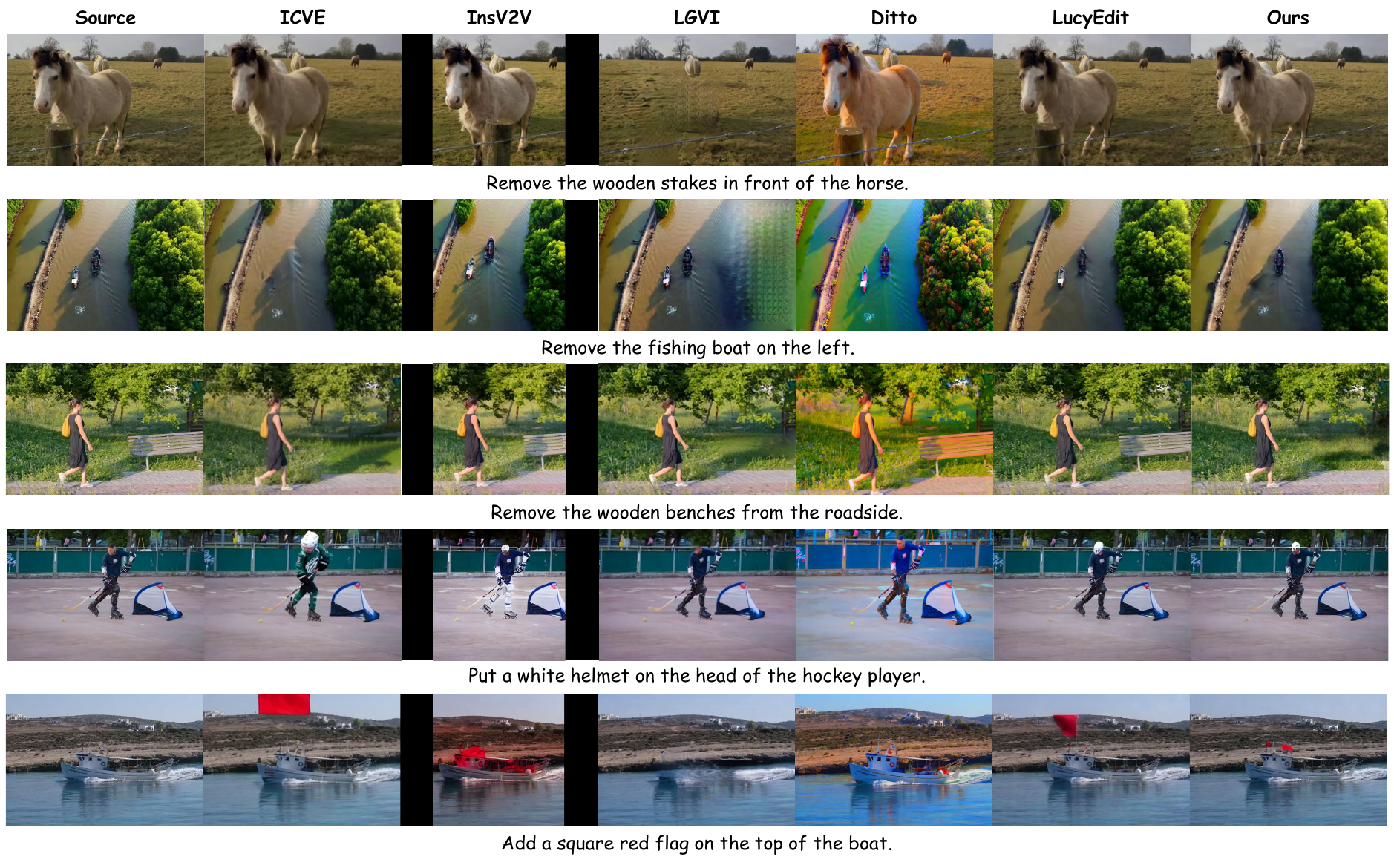}
    \caption{Qualitative comparison on object removal and addition tasks. Each row presents input videos and the corresponding editing results produced by different methods. In contrast, LoVoRA accurately localizes the target regions, cleanly removes or seamlessly inserts objects, and preserves the original background content with stable temporal coherence. 
   }
    \label{fig:qualitative}
\end{figure*}

\section{Experiments}
\label{sec:experiments}
\myparagraph{Training Details.}
Our LoVoRA model is trained on our proposed dataset containing more than 8k video editing pairs. We fine-tune the 3D DiT backbone using LoRA with rank~128, while keeping the VAE and text encoder frozen to ensure stable visual and textual features. The model is optimized using AdamW with random height-width augmentation.  Input video resolutions are randomly sampled between $512\times512$ and $1024\times1024$ to enhance scale robustness. 



\myparagraph{Baseline methods.}
To evaluate the effectiveness of LoVoRA, we compare
our model against representative state-of-the-art diffusion-based video editing methods on both object addition and removal tasks, including Senorita~\cite{senoritia}, ICVE~\cite{icve}, InsV2V~\cite{insv2v}, LGVI~\cite{lgvi2023}, Ditto~\cite{ditto}, and LucyEdit~\cite{lucyedit}. 

\myparagraph{Evaluation Metrics.}
We evaluate model performance from three complementary perspectives: (1) Text Alignment, (2) Video Quality, and (3) VLM assessment. Frame and video Text Alignment\textbf{ \cite{ju2025editverse} }measures how well the generated video matches the editing instruction at both frame and video levels, based on multimodal embedding similarity between text and generated visual content. Video Quality is assessed using four perceptual dimensions from the VBench protocol~\cite{vbench}: subject consistency, background consistency, motion smoothness, aesthetic quality, and imaging quality. These metrics are particularly relevant for object addition and removal tasks. VLM Evaluation further provides high-level semantic assessments using the multimodal large language model MiniCPM-V2.6~\cite{minicpm}. 
It produces structured scores across Prompt Following (PF) and Edit Quality (EQ), thereby offering a holistic view of the model’s instruction understanding and editing realism. 

\myparagraph{Benchmarks.}
We evaluate LoVoRA on a unified benchmark combining DAVIS~\cite{davis} and our LoVoRA-Bench for comprehensive comparison. First, the public DAVIS ~\cite{davis} benchmark provides paired reference frames and original videos, enabling direct assessment of temporal coherence and edit localization accuracy. Second, we construct a dedicated LoVoRA-Bench derived from our dataset generation pipeline, containing curated test cases covering various editing scenarios (e.g., static scenes, small moving objects, occlusions, and long-range motion). 
Editing instructions are automatically generated using InternVL3 ~\cite{InternVL3} to ensure semantic alignment with the visual content. 

\subsection{Quantitative and Qualitative Comparisons}
\label{sec:quantitative_results}

We evaluate LoVoRA and baseline methods on both object removal and object addition tasks. 
Quantitative results are summarized in Tab.~\ref{tab:removal} and~\ref{tab:addition}, while visual comparisons are illustrated in Fig.~\ref{fig:qualitative}. 
Across all metrics and qualitative observations, LoVoRA demonstrates strong spatio-temporal consistency and semantic precision in localized editing.

\myparagraph{Object Removal.}
Tab.~\ref{tab:removal} presents comparison results on the removal task. 
LoVoRA achieves the best or second-best performance across almost all metrics, particularly in text alignment and edit quality. Senorita~\cite{senoritia} benefits from using reference images during inference, which explains its relatively high subject consistency. ICVE~\cite{icve} produces reasonable visual quality but its localization of editing positions can be further improved, especially for small or partially occluded targets. Compared with Ditto~\cite{ditto}, which performs video retouching and thus enhances aesthetic quality but weakens local edit accuracy, our model maintains precise object removal without over-smoothing the scene.
LucyEdit~\cite{lucyedit} shows limited ability in removing target objects, often leaving visible remnants or distortions near boundaries. LGVI~\cite{lgvi2023} tends to over-remove contextual elements. 
InsV2V~\cite{insv2v} only supports square-cropped videos, which restricts its applicability to general high-resolution video inputs. 
As shown in Fig.~\ref{fig:qualitative}, LoVoRA preserves background integrity while accurately removing objects and maintaining consistent lighting and motion cues across frames.

\myparagraph{Object Addition.}
Tab.~\ref{tab:addition} reports the results for the object addition task. 
LoVoRA again delivers the most balanced performance across all metrics, producing high-quality inserted objects while maintaining background integrity and temporal consistency. Senorita~\cite{senoritia} achieves the highest video-text alignment scores in some cases, which can be attributed to its use of reference images to guide the insertion. LucyEdit~\cite{lucyedit} performs reasonably well in addition scenarios, producing plausible insertions but occasionally introducing boundary blur. Ditto’s~\cite{ditto} editing design improves overall aesthetics but fails to perform localized insertions faithfully. For ICVE~\cite{icve}, the added objects may appear misplaced or contextually inconsistent, and this issue becomes more pronounced when inserting small or thin objects. LGVI~\cite{lgvi2023} struggles with over-editing and temporal drift, while InsV2V~\cite{insv2v} often modifies unrelated background regions during addition. The qualitative results in Fig.~\ref{fig:qualitative} further confirm that LoVoRA generates seamlessly integrated objects with consistent color tone, motion coherence, and minimal background distortion.


\subsection{Ablation Study}
\label{sec:ablation}

\begin{figure}[t]
    \centering
    \setlength{\abovecaptionskip}{2pt}
    \includegraphics[width=1\linewidth]{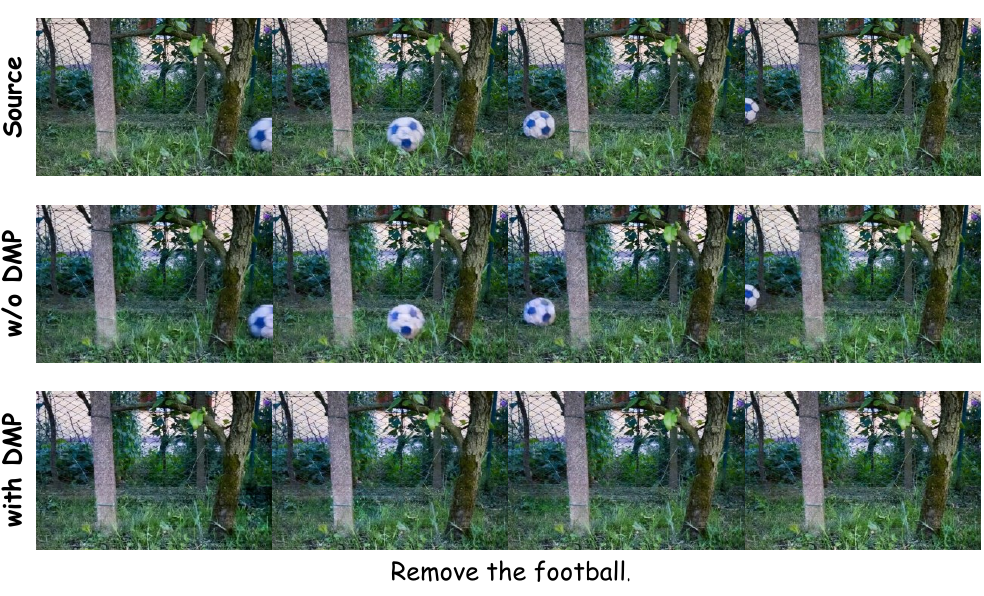}
    \caption{Ablation study on the Diffusion Mask Predictor (DMP). 
    Compared to the model trained without DMP, LoVoRA with DMP produces more accurate localization.}
    \vspace{-0.8em}
    \label{fig:ablation}
\end{figure}

To validate the contribution of the proposed Diffusion Mask Predictor (DMP), 
we perform an ablation study by training LoVoRA with and without this module under identical settings.
As shown in Fig.~\ref{fig:ablation}, 
without DMP, the model struggles to accurately localize edited regions. By contrast, incorporating DMP yields more precise spatial boundaries and improved temporal stability. This improvement stems from the DMP’s ability to learn soft spatio-temporal attention masks, guiding the diffusion process to focus on semantically relevant areas. These results demonstrate that the Diffusion Mask Predictor is essential for achieving reliable text-guided mask-free object-level video editing.
Additional ablation results on dataset construction are provided in the supplementary material.

\section{Conclusion}
\label{sec:conclusion}
In this paper, we introduced LoVoRA, a text-guided and mask-free framework for video object removal and addition. Unlike prior approaches that rely on user-provided masks or reference signals, LoVoRA achieves fully end-to-end editing by learning an object-aware localization mechanism that implicitly identifies and modifies target regions based solely on text instructions. Supported by a newly constructed video editing dataset featuring temporally consistent optical-flow supervision and semantically aligned textual guidance, our framework enables fine-grained, motion-aware video manipulation without manual intervention. Through extensive experiments and human evaluation, LoVoRA demonstrates superior performance in spatial precision, temporal stability, and semantic alignment compared with recent instruction-based video editing models. Our findings validate that integrating learnable localization with diffusion-driven generation offers a scalable and generalizable solution for text-based video editing. These results suggest that learning implicit localization from text provides a promising direction for achieving general and scalable video editing without external control signals.

\newpage
{
    \small
    \bibliographystyle{ieeenat_fullname}
    \bibliography{main}
}

\clearpage
\setcounter{page}{1}
\maketitlesupplementary

\section{Ablation Study on Dataset}
\label{sec:dataset_ablation}

To further validate the effectiveness of the proposed LoVoRA dataset, we conduct a series of dataset ablation experiments. Our goal is to assess how different data sources influence the learning of mask-free object-aware localization and the final video editing performance. We evaluate four training configurations:
\begin{enumerate}
    \item \textbf{Senorita-2M Only:} Training LoVoRA solely on Senorita-2M~\cite{senoritia}.
    \item \textbf{Ditto-1M Only:} Training LoVoRA solely on Ditto-1M~\cite{ditto}.
    \item \textbf{Ours Only:} Training LoVoRA on our synthesized dataset.
    \item \textbf{Ours + Senorita (Mixed 50\% : 50\%):} Joint training using an equal mixture of our dataset and Senorita-2M.
\end{enumerate}

All models share identical architectures, training schedules, and evaluation protocols. We evaluate on the unified benchmark combining LoVoRA-Bench and DAVIS~\cite{davis} on removal and addition tasks. The metrics are the same as defined in the main paper: Text Alignment (Frame / Video), Video Quality (Subject Consistency, Background Consistency, Motion Smoothness, Aesthetic Quality, Imaging Quality), and VLM-based Prompt Following (PF) / Edit Quality (EQ).

\subsection{Quantitative Comparison}
Tab.~\ref{tab:data_ablation} summarizes the results. Training solely on Senorita-2M~\cite{senoritia} or Ditto-1M~\cite{ditto} yields reasonable global video quality but the model receives only weak and spatially ambiguous editing signals without explicit mask supervision. As a result, it struggles to learn reliable object-level localization. Our dataset provides dense, flow-aligned spatio-temporal cues that clarify where edits should occur throughout a video, enabling the model to acquire stronger object-aware reasoning. This leads to clear improvements in both text alignment and visual fidelity. The mixed dataset further strengthens performance by combining Senorita’s scene diversity with our precise localization supervision, resulting in the most robust and balanced editing quality.

\begin{table*}[t]
\centering
\caption{Quantitative comparison on the object addition task.
We report results across three evaluation dimensions: Text Alignment, Video Quality, and VLM-based evaluation. PF denotes Prompt Following and EQ denotes Edit Quality. Best results are bolded.}
\label{tab:data_ablation}
\resizebox{\linewidth}{!}{
\begin{tabular}{c|cc|ccccc|cc}
\toprule
\multirow{2}{*}{\textbf{Dataset}} &
\multicolumn{2}{c|}{\textbf{Text Alignment}$\uparrow$} &
\multicolumn{5}{c|}{\textbf{Video Quality}$\uparrow$} &
\multicolumn{2}{c}{\textbf{VLM}$\uparrow$} \\
\cmidrule(lr){2-3} \cmidrule(lr){4-8} \cmidrule(lr){9-10}
&  Frame & Video & Subj.\ Cons. & Backg.\ Cons. & Motion & Aesthetic & Imaging & PF & EQ \\
\midrule
Senorita-2M Only & 19.57 & 14.83 & 0.9084 & 0.9019 & 0.9806 & 0.5366 & 0.5941 & 3.913 & 4.510 \\
Ditto-1M Only     & 21.39 & 15.89 & 0.9124  & 0.8993 & 0.9826 & 0.5627 & 0.6150 & 3.539 & 4.216 \\
Ours Only         & 21.82 & \textbf{16.42} & \textbf{0.9315} & 0.9156 & 0.9874 & 0.5575 & 0.6286 & \textbf{4.129} & 4.548 \\
Ours + Senorita   & \textbf{21.96} & 16.20 & 0.9310 & \textbf{0.9172} & \textbf{0.9887} & \textbf{0.5680} & \textbf{0.6412} & 3.955 & \textbf{4.609} \\
\bottomrule
\end{tabular}}
\end{table*}

\subsection{Qualitative Comparison}
\begin{figure*}[t]
\centering
\includegraphics[width=\textwidth]{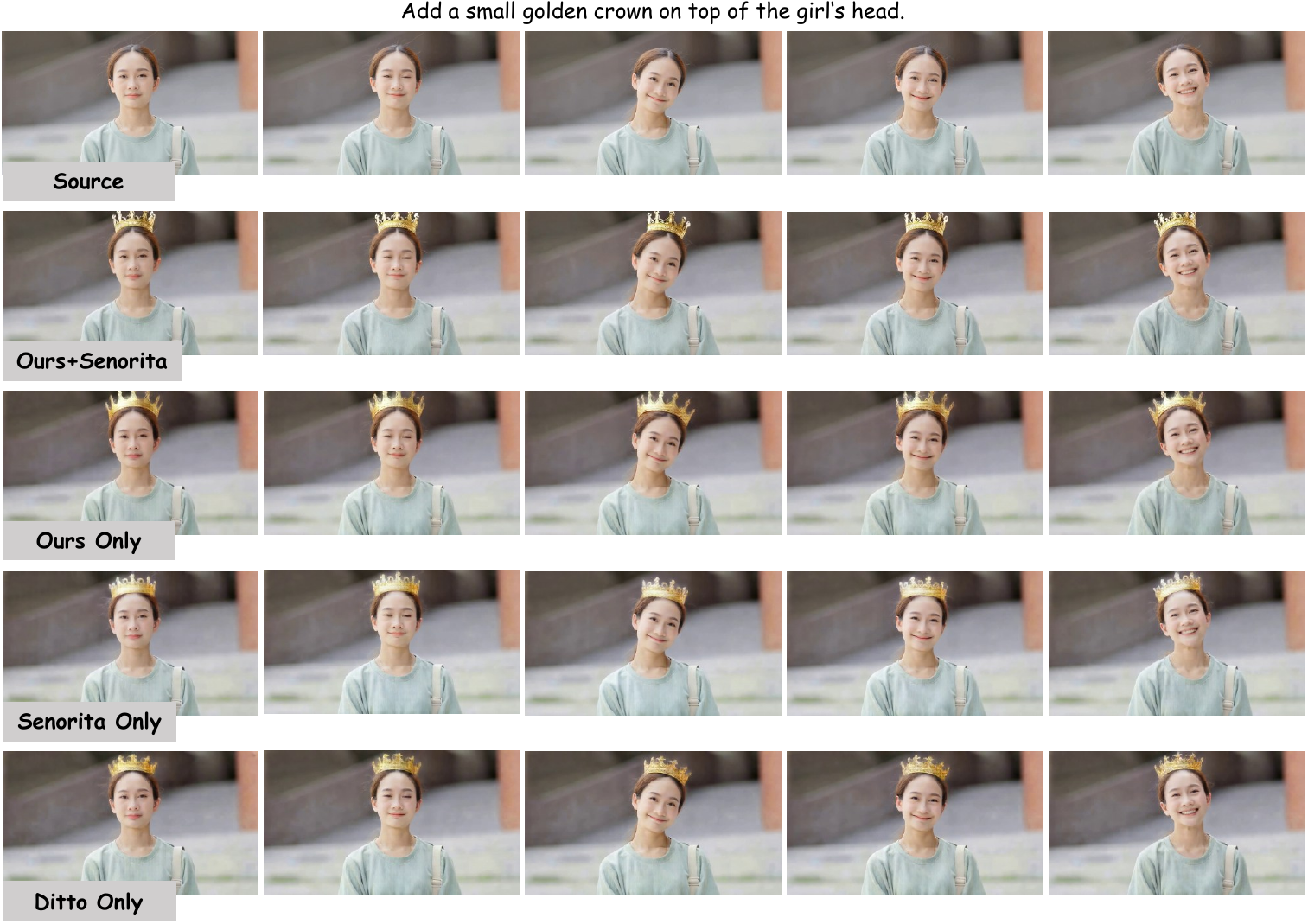}
\caption{Dataset ablation: training on our dataset yields sharper localization and stronger temporal consistency, while mixing with Senorita-2M achieves the best overall performance.}
\label{fig:data_ablation}
\end{figure*}

Fig.~\ref{fig:data_ablation} illustrates visual comparisons. Models trained on Senorita-2M ~\cite{senoritia} and Ditto-1M ~\cite{ditto} often struggle with object boundary precision and introduce temporal flickering. In contrast, our dataset’s flow-propagated masks provide clear, temporally aligned editing cues, enabling the model to preserve object shape, position, and motion consistency even in challenging scenarios. The mixed dataset further improves stability in complex motion scenarios, confirming that our optical-flow-guided mask propagation provides strong supervision for object-aware localization.

Although Senorita-2M~\cite{senoritia} and Ditto-1M~\cite{ditto} provide large-scale edited videos, their edit operations are either globally applied or lack explicit object-level motion annotations. This introduces ambiguity in the mapping between the textual instruction and the corresponding edited region. Such ambiguity directly harms the model’s ability to learn object-specific modifications, leading to drifting boundaries, content hallucination, and unstable temporal behaviors. Our dataset resolves this fundamental issue by explicitly embedding spatio-temporal editing supervision in the training pairs.


\section{Additional Experiments}
\label{sec:additional_experiments}


\label{subsec:more_methods}

In addition to the text-based baselines evaluated in the main paper, we conduct further comparisons against several widely used video editing frameworks:

\begin{itemize}
    \item \textbf{AnyV2V}~\cite{ku2024anyv2v} (reference-guided)
    \item \textbf{Senorita-2M}~\cite{senoritia} (reference-guided)
    \item \textbf{ICVE}~\cite{icve} (instruction-based)
    \item \textbf{Kling} (commercial)
\end{itemize}

As these tools provide reference images during inference, we conduct a \textbf{user study} to compare their editing effectiveness against LoVoRA. Participants rated each method under two criteria:
(1) Editing Completeness: How well the video follows the target editing instruction. (2) Video Quality: Temporal consistency, realism, and lack of artifacts.

Tab.~\ref{tab:user_study} shows that LoVoRA achieves strong editing completeness and video quality despite receiving no reference images, outperforming all reference-based methods and approaching the fidelity of commercial systems. Notably, Kling achieved the highest subjective scores, which is attributed to its global enhancement and commercial-level video restoration. However, participants also reported that its outputs can deviate from strict object-level instructions, such as adding or reshaping unrelated background details. In contrast, LoVoRA preserves user intent more faithfully: it performs targeted object manipulations while keeping unrelated regions intact, leading to a more controllable and semantically reliable editing experience. 

\begin{table}[h]
\centering
\caption{User study results comparing LoVoRA with reference-based, instruction-based, and commercial video editing systems on Editing Completeness (EC) and Video Quality (VQ). Scores are averaged across all evaluation videos (1–5 scale; higher is better).}
\label{tab:user_study}
\begin{tabular}{l|cc}
\toprule
Method & EC↑ & VQ↑ \\
\midrule
AnyV2V ~\cite{ku2024anyv2v}      & 2.875 & 2.416 \\
Senorita ~\cite{senoritia}    & 3.492 & 2.917 \\
ICVE ~\cite{icve} & 2.925 & 2.742 \\
Kling & \textbf{4.016} & \textbf{3.733} \\
\midrule
\textbf{LoVoRA (ours)} & \underline{3.675} & \underline{3.283} \\
\bottomrule
\end{tabular}
\end{table}

\begin{figure*}[t]
\centering
\includegraphics[width=\textwidth]{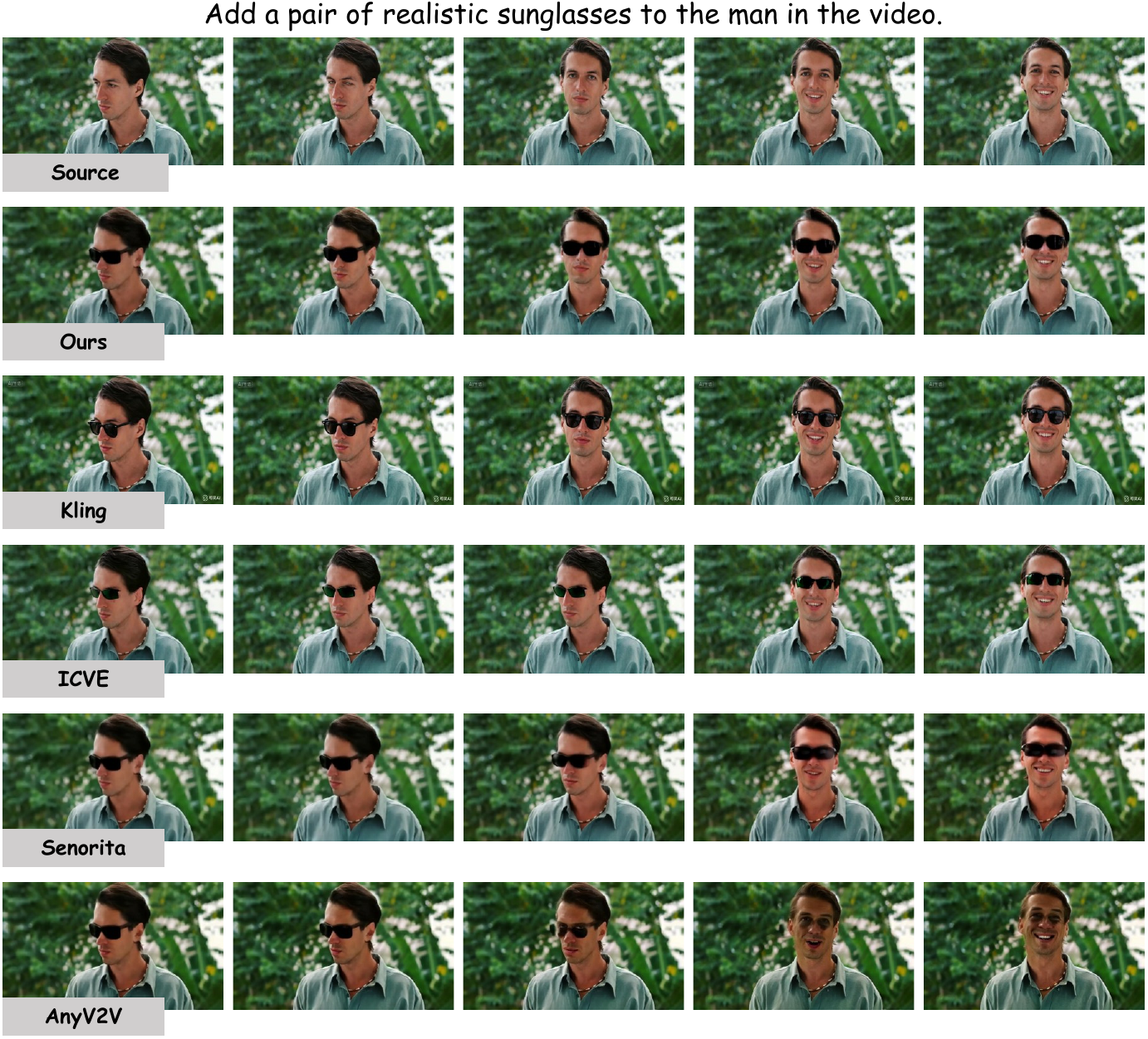}
\caption{Visual comparisons with more methods. LoVoRA achieves strong semantic alignment and stable editing without requiring references.}
\label{fig:man}
\end{figure*}

\begin{figure*}[t]
\centering
\includegraphics[width=\textwidth]{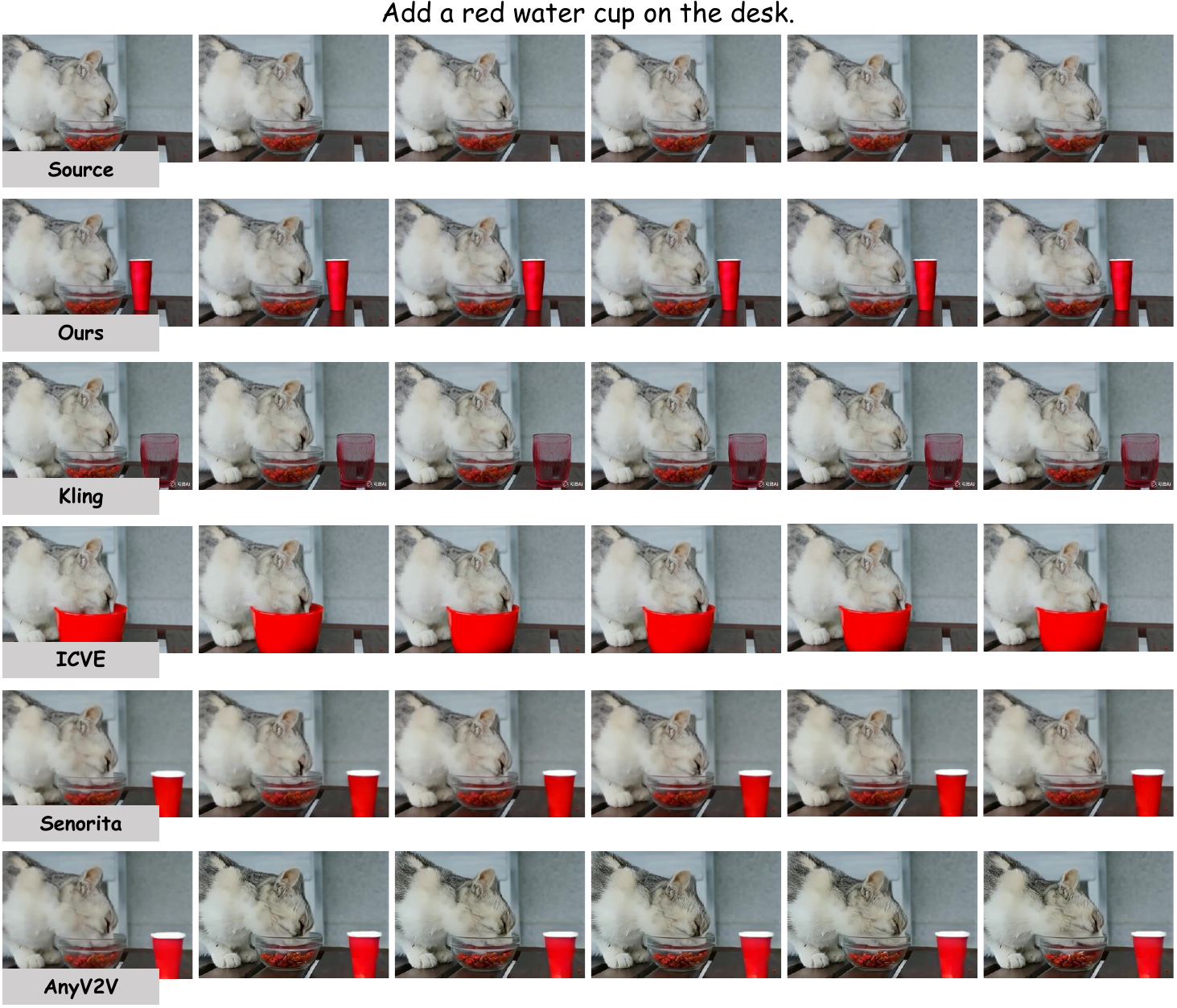}
\caption{Visual comparisons with more methods. LoVoRA achieves strong semantic alignment and stable editing without requiring references.}
\label{fig:cat}
\end{figure*}

\begin{figure*}[t]
\centering
\includegraphics[width=\textwidth]{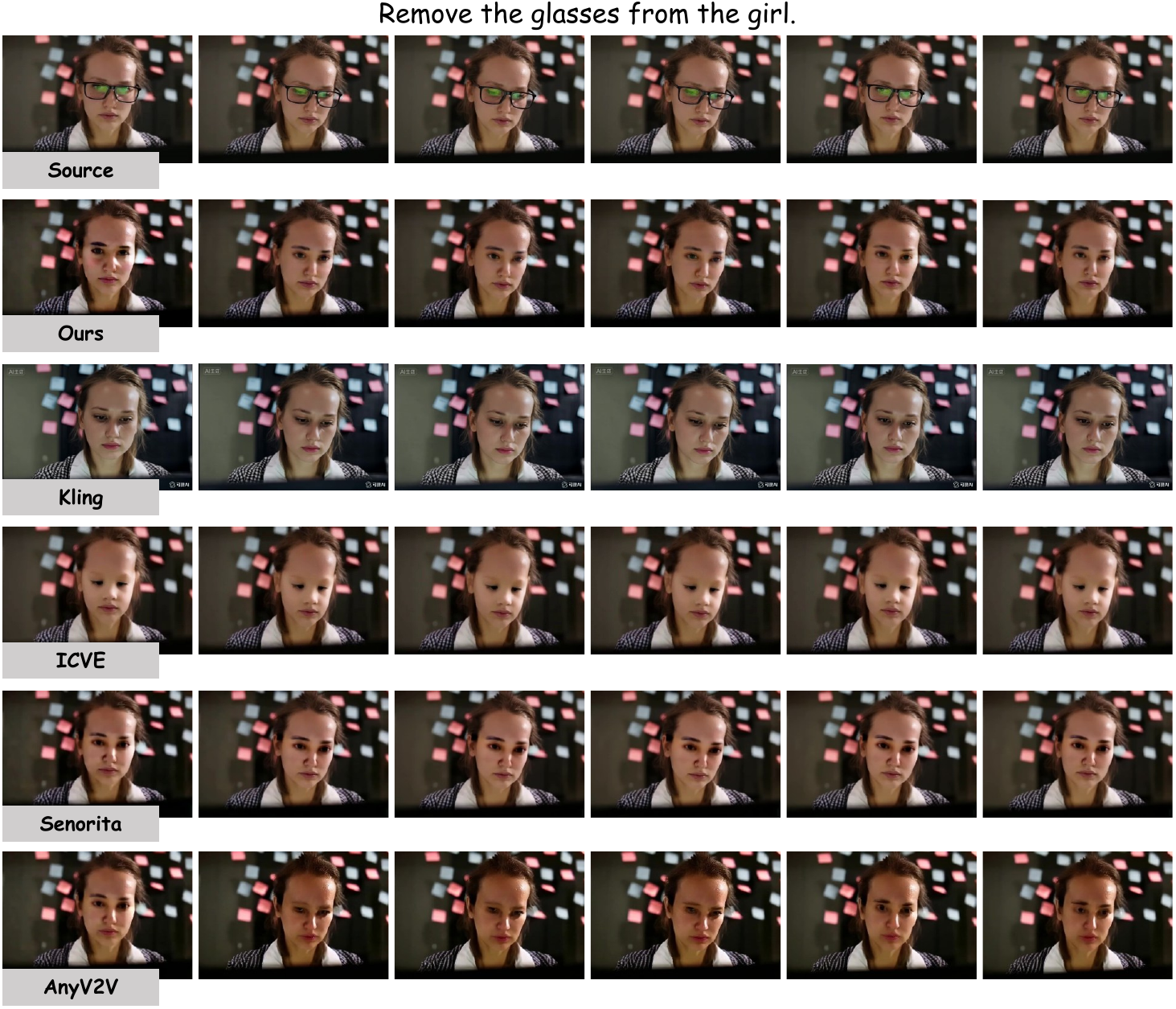}
\caption{Visual comparisons with more methods. LoVoRA achieves strong semantic alignment and stable editing without requiring references.}
\label{fig:glass}
\end{figure*}

\begin{figure*}[t]
\centering
\includegraphics[width=\textwidth]{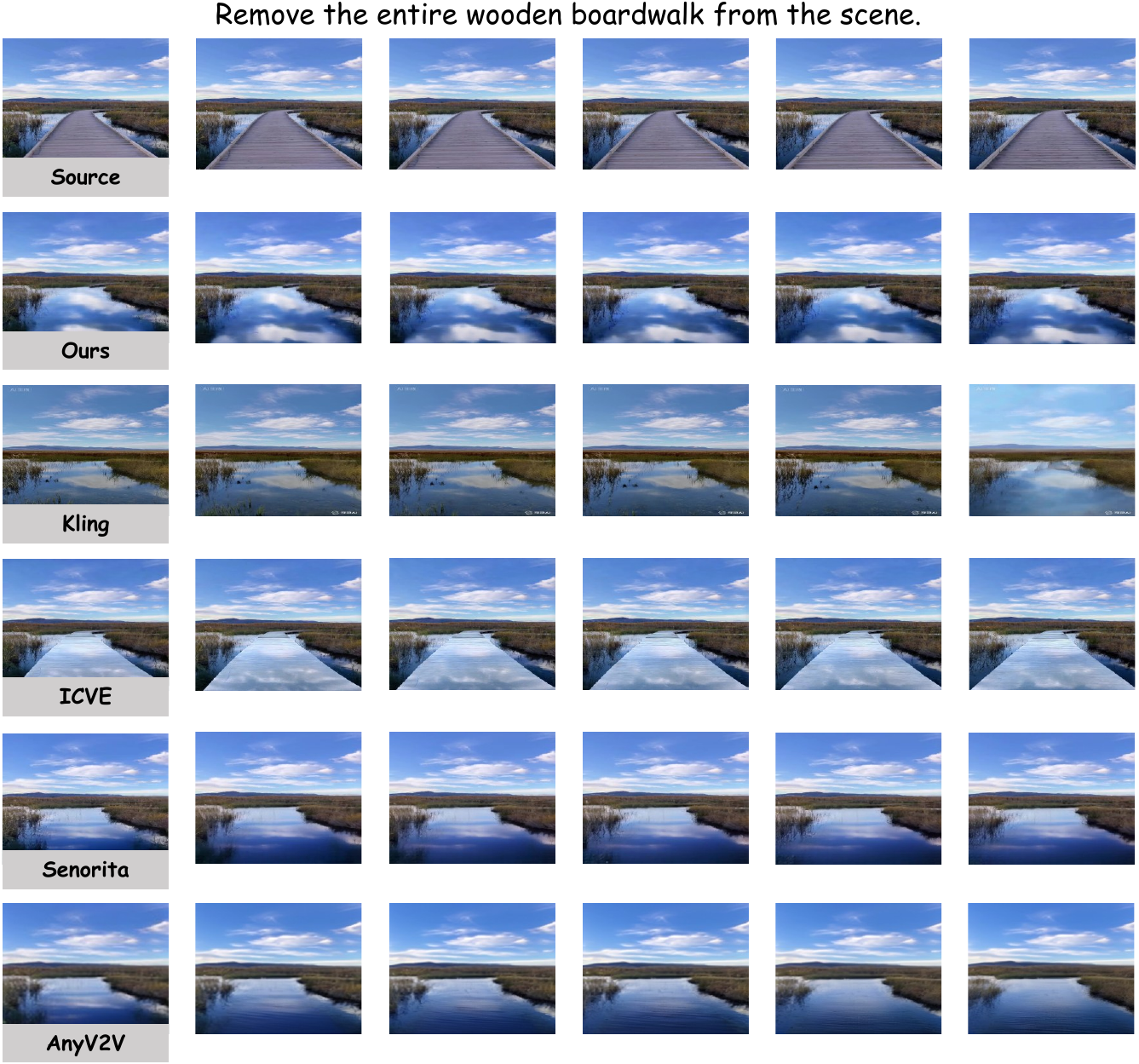}
\caption{Visual comparisons with more methods. LoVoRA achieves strong semantic alignment and stable editing without requiring references.}
\label{fig:remove}
\end{figure*}

Fig.~\ref{fig:man}, Fig.~\ref{fig:cat}, Fig.~\ref{fig:glass}, and Fig.~\ref{fig:remove} provide diverse qualitative comparisons across removal and addition scenarios. Reference-based frameworks typically render sharp textures but tend to drift from the text instruction, often altering object types, colors, or spatial layout due to dependency on reference-image similarity rather than textual semantics. Kling exhibits another pattern where its global enhancement produces clean, high-quality videos but frequently modifies background content beyond the local editing region. This occurs especially when the original footage contains motion blur or low-quality frames. This raises an interesting ambiguity in commercial models—--whether fidelity to local editing or global visual optimization should take priority. LoVoRA, by contrast, consistently restricts edits to the intended object regions and maintains temporal coherence without altering unrelated areas, demonstrating stable behavior across all editing types and input qualities.

\section{Diverse Instructions}
\label{subsec:change}

We further evaluate LoVoRA under object-level replacement instructions. These instructions require localized removal of an existing object followed by the synthesis of a new one, posing stronger challenges in spatial reasoning, temporal coherence, and region-specific appearance preservation.

\begin{figure*}[t]
\centering
\includegraphics[width=\textwidth]{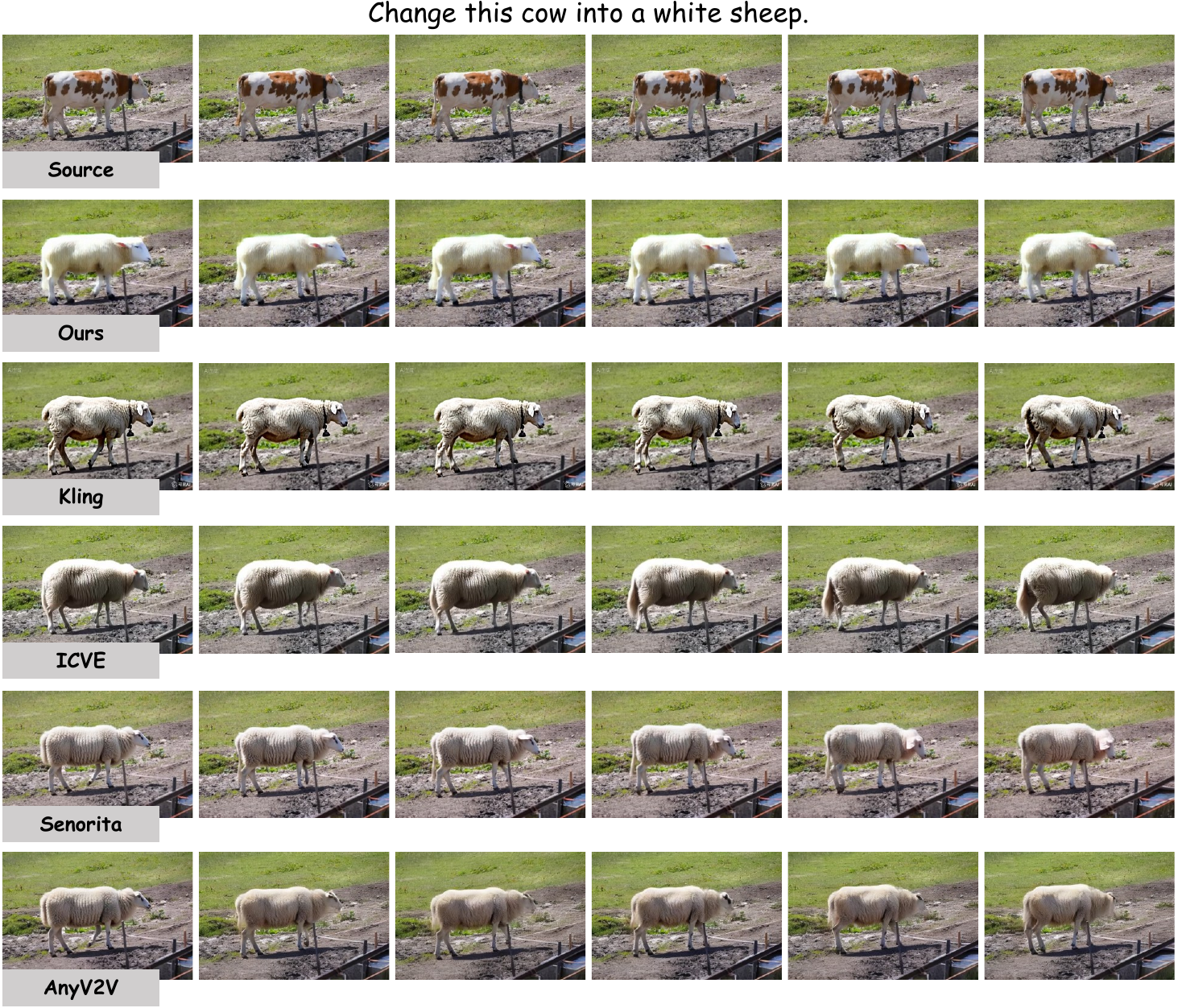}
\caption{LoVoRA performance on object-level replacement instructions. The model removes the source object and synthesizes the target object with stable motion and appearance.}
\label{fig:change_cases}
\end{figure*}

Fig.~\ref{fig:change_cases} demonstrates that LoVoRA can handle such replacement scenarios reasonably well. The model accurately removes the original object, synthesizes the new content, and preserves consistent motion and lighting across frames. Since change editing inherently combines removal and addition, our unified mask-free localization mechanism easily generalizes to these scenarios.

We extend our dataset pipeline to support object replacement instructions. In this case, we extract two masks from $(I_s, I_t)$ representing the removed and inserted objects, respectively.
The initial mask is defined as the union:
\begin{equation}
M_1 = m^{(s)} \lor m^{(t)}.
\end{equation}
The union mask is then refined and propagated through the optical flow field to form $\{M_t\}_{t=1}^{T}$. This ensures that both disappearing and newly synthesized content are supervised consistently across time. The inpainting stage subsequently produces $V_t$ conditioned on the unified mask flow, allowing the dataset to naturally support replacement edits.

\end{document}